\documentclass[letterpaper, 10 pt, conference]{ieeeconf}  

\IEEEoverridecommandlockouts                              

\usepackage{graphics} 
\usepackage{epsfig} 
\usepackage{mathptmx} 
\usepackage{times} 
\usepackage{amsmath} 
\usepackage{amssymb}  
\usepackage[ruled,linesnumbered]{algorithm2e} 
\usepackage{cite} 
\usepackage{url} 
\usepackage{booktabs}    
\usepackage{multirow} 
\usepackage{bm}

\title{\LARGE \bf
HEATS: A Hierarchical Framework for Efficient Autonomous Target Search with Mobile Manipulators
}

\author{Hao Zhang$^{1}$, Yifei Wang$^{1}$, Weifan Zhang$^{1}$, Yu Wang$^{2}$ and Haoyao Chen$^{1,*}$
\thanks{*Corresponding author.}
\thanks{$^{1}$H. Zhang, Y.F. Wang, W.F. Zhang and H.Y. Chen are with the School of Mechanical Engineering and Automation, Harbin Institute of Technology Shenzhen, P.R. China.
        {\tt\small x1ezhhter@gmail.com, \{wangyifei, 23S153018\}@stu.hit.edu.cn, hychen5@hit.edu.cn}.}%
\thanks{$^{2}$Y. Wang is with the Department of Precision Mechinery and Precision Instrumentation, University of Science and Technology of China.
        {\tt\small wangyuustc@ustc.edu.cn}.}%
\thanks{$^{3}$\protect\url{https://github.com/Andy168byte/HEATS}.}
}

\begin{document}

\maketitle
\thispagestyle{empty}
\pagestyle{empty}

\begin{abstract}

Utilizing robots for autonomous target search in complex and unknown environments can greatly improve the efficiency of search and rescue missions.
However, existing methods have shown inadequate performance due to hardware platform limitations, inefficient viewpoint selection strategies, and conservative motion planning.  
In this work, we propose HEATS, which enhances the search capability of mobile manipulators in complex and unknown environments.
We design a target viewpoint planner tailored to the strengths of mobile manipulators, ensuring efficient and comprehensive viewpoint planning.  
Supported by this, a whole-body motion planner integrates global path search with local IPC optimization, 
enabling the mobile manipulator to safely and agilely visit target viewpoints, significantly improving search performance. 
We present extensive simulated and real-world tests, in which our method demonstrates reduced search time, 
higher target search completeness, and lower movement cost compared to classic and state-of-the-art approaches.  
Our method will be open-sourced for community benefit$^{3}$.

\end{abstract}

\begin{keywords}
    Reactive and Sensor-Based Planning, Search and Rescue Robots, Motion and Path Planning, Field Robots.
\end{keywords}

\section{Introduction}

Tasks such as explosive ordnance disposal (EOD) \cite{9570974}, rescue \cite{Wirth_Pellenz_2007}, and contaminant remediation \cite{Yokokohji_2021} 
necessitate precise target localization in unknown environments.
Autonomous robots can significantly reduce human exposure time in hazardous scenarios by independently conducting target search missions. 
During the search process, robots must not only efficiently explore unknown regions 
but also conduct visual inspection of suspicious areas to locate targets, seamlessly integrating exploration with inspection.

Conventional mobile robots (UAVs/UGVs) equipped with onboard sensors such as LiDAR and cameras can 
effectively perform target search tasks in unknown environments \cite{10166021}\cite{10451197}\cite{Star}. 
However, in narrow and occluded scenarios, their exploration completeness and efficiency are significantly compromised by size constraints and fixed sensor perspectives. 
Mobile manipulators overcome visual occlusions by leveraging their high degree of freedom (DoF) to generate optimal viewpoints for sensors mounted on the end-effector. 
Nevertheless, existing algorithms fail to fully exploit these advantages.
First, current Next-Best-View (NBV) generation strategies neglect the high-DoF characteristics of mobile manipulators, 
resulting in insufficient search completeness. 
Second, the prevalent decoupled planning approach leads to overly conservative motion trajectories, thereby hindering search efficiency.

\begin{figure}[t]
    \centering
    \includegraphics[scale=0.2]{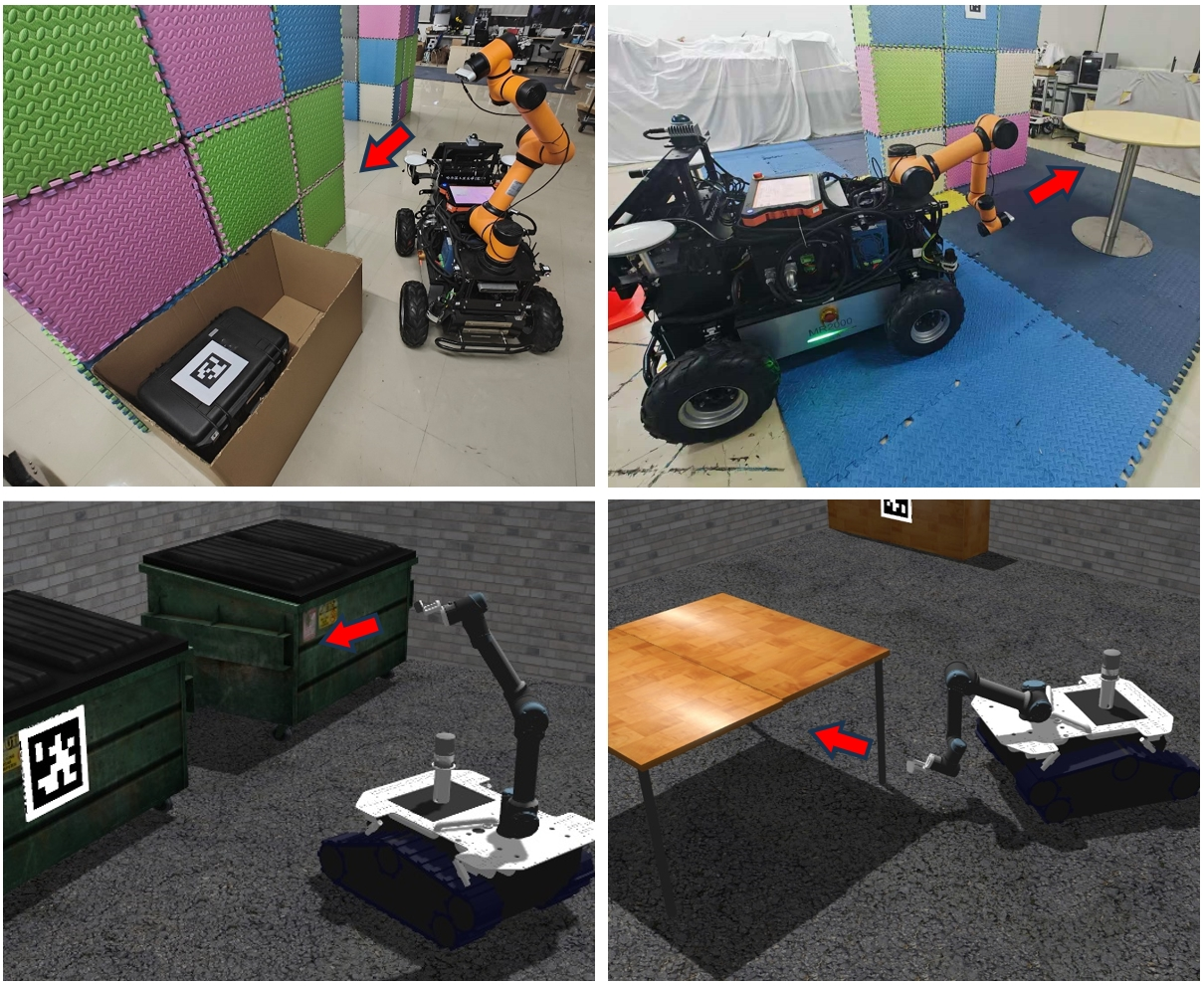}
    \caption{The mobile manipulator searches for targets (represented by apriltags) in both real indoor environments and simulation environments.
    The target search completeness is enhanced by leveraging the high degree of freedom of the mobile manipulator, 
    allowing the camera at the end-effector to cover all areas of the scene.}
    \label{Intro}
\end{figure}

\begin{figure*}[t]
    \centering
    \includegraphics[scale=0.255]{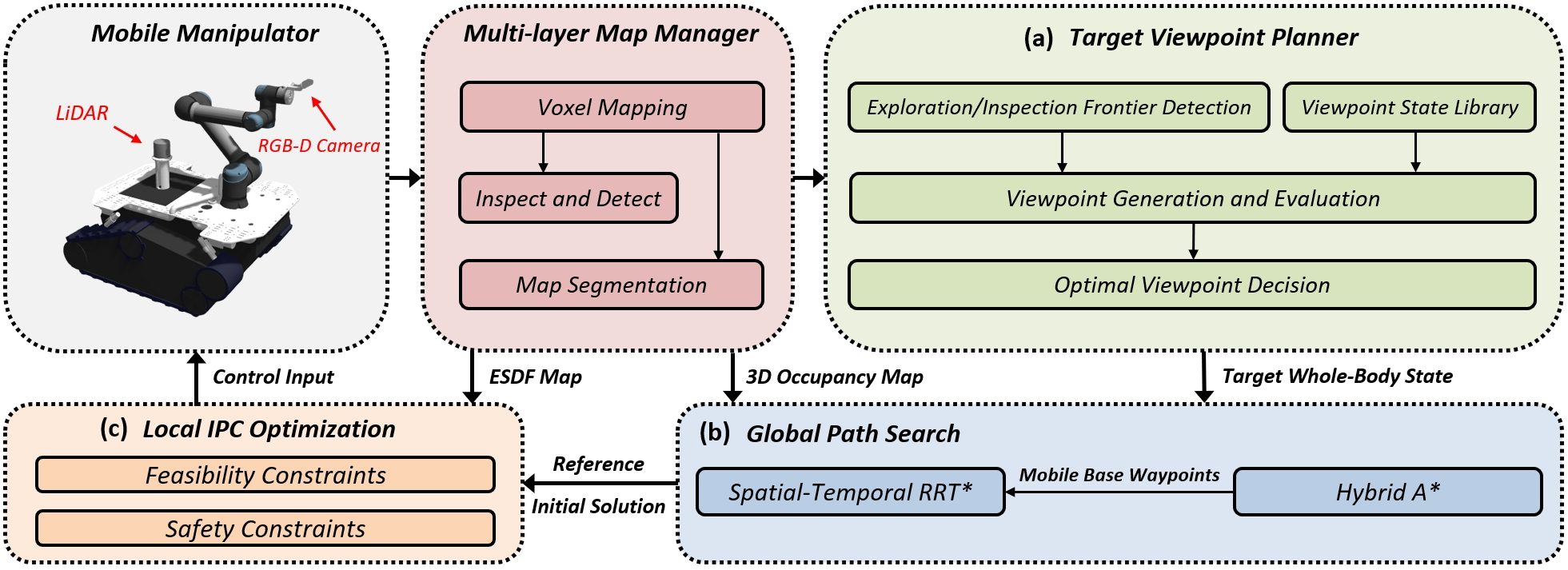}
    \caption{The overview of the proposed hierarchical framework for efficient autonomous target search.}
    \label{system}
\end{figure*}

To address these challenges, we propose \textbf{HEATS}, a \textbf{H}ierarchical framework designed for mobile manipulators 
to achieve \textbf{E}fficient \textbf{A}utonomous \textbf{T}arget \textbf{S}earch in complex and unknown environments.
We introduce a target viewpoint planner that determines optimal viewpoints for the mobile manipulator in real-time. 
The algorithm first generates and evaluates candidate viewpoints through a viewpoint state library and a well-designed utility function. 
Then a dual-stage decision module is employed to select the optimal viewpoint, balancing search efficiency and completeness.                                                                                                                                                                                                                                                                   
To enable agile and safe visiting of viewpoints in complex environments, our framework incorporates a global path search module
to generate a high-quality whole-body initial path for the mobile manipulator. 
An Integrated Planning and Control (IPC) optimization module further enhances motion safety and trajectory smoothness through certain constraints.

We compared our approach to state-of-the-art and classic methods in simulation. The results show that our method achieves superior target search performance in all experiments. 
We further validated our system in complex real-world environments using entirely onboard devices.
The contributions of this paper are summarized as follows:

\begin{itemize}

\item A target viewpoint planner designed for mobile manipulators, which introduces a viewpoint state library and a dual-stage decision module to determine the optimal viewpoint, ensuring comprehensive autonomous target search.
\item A whole-body motion planner that integrates global path search with IPC optimization, enabling robots to navigate complex environments agilely and safely while visiting optimal viewpoints.
\item Extensive simulation and real-word tests that validate the proposed method. The source code of our system will be made public.

\end{itemize}

\section{Related Works}

Autonomous target search requires robots to simultaneously perform exploration and visual inspection in unknown environments.
Most solutions for exploration tasks adopt either frontier-based methods or sampling-based methods.
Frontier-based methods define frontiers as the boundary regions between known and unknown areas and determine the robot's target points by calculating the information gain \cite{frontier_1}\cite{frontier_2}. 
Some algorithms further enhance global optimality by formulating the Traveling Salesman Problem (TSP) to assign visitation sequences to frontiers \cite{frontier_3}\cite{tare}.
Sampling-based methods generate a tree-like candidate set of viewpoints through sampling strategies 
and select the branch with the highest information gain as the robot's exploration path \cite{Schmid_Pantic_Khanna_Ott_Siegwart_Nieto_2020}\cite{sample_2}.
Additionally, some approaches combine the strengths of both methods by first planning a global path to frontiers and then sampling paths locally\cite{COME1}. 

Several studies have integrated exploration tasks with other mission-specific objectives. 
Papatheodorou et al. \cite{Semantic} designed a utility function for target searching and objects reconstruction in unknown environments.
Gao et al. \cite{Trolley} proposed a task-oriented environment partitioning algorithm, enabling robots to collect specific objects within rooms while performing exploration. 
Luo et al. \cite{Star} incorporated historical information into the global decision-making module for target search tasks, enhancing path consistency. 

Algorithms designed for UAVs/UGVs cannot be directly applied to mobile manipulators due to kinematic and task constraints. 
Consequently, exploration and inspection algorithms specifically tailored for mobile manipulators have gained increasing attention.  
Isler et al. \cite{Pin} proposed an information-gain-based viewpoint sampling strategy, 
which drives the manipulator's end-effector to perform periodic motions in predefined patterns to expand the perceptual range. 
Menaka et al. \cite{Sota} allocated task-specific weights within the information gain function, enabling the robot to explore the environment and inspect highly contaminated areas. 
However, these methods fail to fully exploit the high DoF advantages of mobile manipulators at the viewpoint planning level, 
while their motion planning strategies tend to be overly conservative, resulting in inefficient and incomplete target search performance. 
In contrast, we introduce a target viewpoint planner designed for mobile manipulators, 
coupled with a safe and efficient motion planner, achieving efficient and comprehensive target search.

\section{System Overview}

We model the environment as a bounded unknown volume $V \subset \mathbb{R}^3$, discretized into cubic voxels. 
By excluding unobservable regions $V_{uno} \subset V$ (e.g. the inside of walls), 
conventional exploration tasks incrementally divide the observable space $V_{o} = V \setminus V_{uno}$ into free space $V_{free} \subset V_{o}$
and occupied space $V_{occ} \subset V_{o}$ through sensor observations. 
For target search missions, the mobile manipulator must simultaneously explore the environment 
and perform thorough visual inspections of $V_{occ}$ to detect spatially distributed targets. 
The task is considered complete when the entire environment is fully explored and all occupied regions ($V_{occ}$) have been inspected.

The proposed framework is illustrated in Fig. \ref{system}. The mobile manipulator is equipped with a depth camera on the end-effector 
and a LiDAR on the mobile base. 
A multi-layer map manager fuses LiDAR point clouds and depth images from sensors to update voxel occupancy probabilities, 
while RGB images are utilized to inspect occupied voxels and detect targets. 
The target viewpoint planner (Sect. IV) then determines the optimal viewpoint based on the updated map in real-time
and provides the corresponding whole-body target state to the motion planner. 
Subsequently, the global path search module (Sect. V-A) generates a feasible, collision-free whole-body path as an initial solution and reference states, 
guiding the local IPC optimization module (Sect. V-B) to compute control policies for efficient and safe target search by the mobile manipulator.

\section{Target Viewpoint Planner}

The target viewpoint planner designed for mobile manipulators is illustrated in Fig. \ref{system}(a).
The algorithm first extracts exploration and inspection frontiers from the map, 
then utilizes a viewpoint state library to generate and evaluate coverage viewpoints, 
and finally constructs a dual-stage decision module to determine the optimal viewpoint based on region partition results.

\begin{figure}[t]
    \centering
    \includegraphics[scale=0.23]{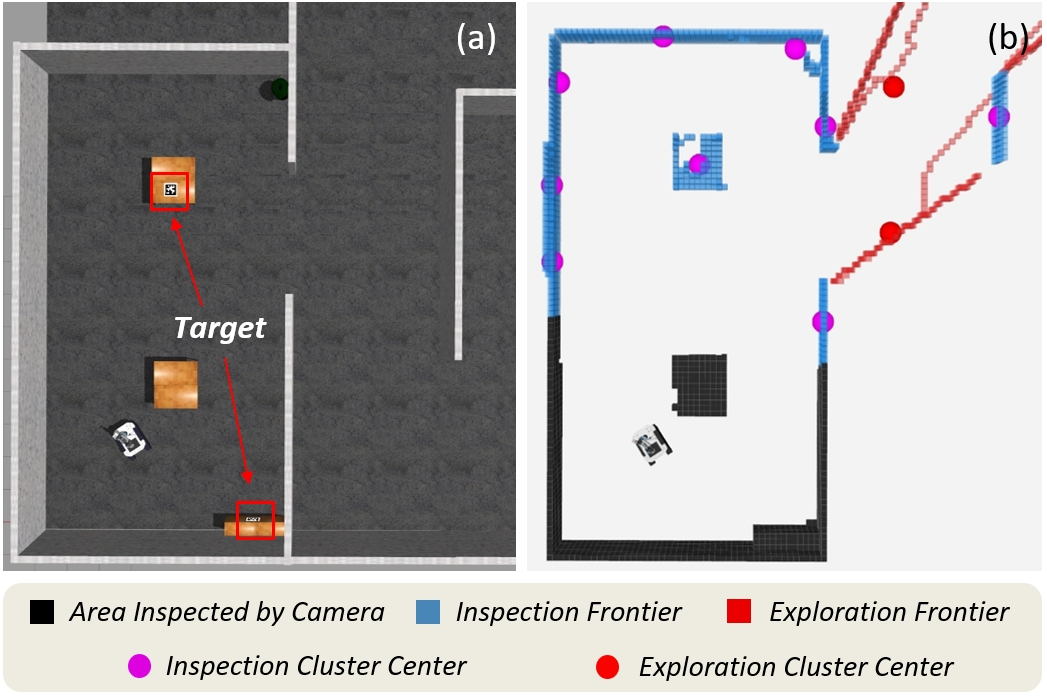}
    \caption{The mobile manipulator detects frontiers in simulation.
    (a) Apriltags are randomly placed within an office environment and used as search targets.
    (b) The robot updates the map information using LiDAR and depth data while inspecting $V_{occ}$. 
    Based on the properties of the voxels, exploration frontiers and inspection frontiers are extracted and clustered.}
    \label{detect}
\end{figure}

\subsection{Exploration and Inspection Frontier Detection} 

As illustrated in Fig. \ref{detect}(b), the map manager updates voxel occupancy probabilities using LiDAR and depth data through ray-casting, dividing unknown space into $V_{free}$ and $V_{occ}$.
Simultaneously, we project the occupied voxels into the camera coordinate frame, calculating the observation distances for 
those within the camera's field of view (FOV) and not occluded by any occupied areas. 
Voxels with the observation distance less than $d_{ins}$ are labeled as $V_{inspected}$ (black regions in Fig. \ref{detect}(b)),
indicating that these voxels have been visually inspected by the camera \cite{Star}.
Building on this, exploration frontiers are defined as free voxels adjacent to unknown voxels (red voxels in Fig. \ref{detect}(b)), while inspection frontiers are occupied voxels neighboring free voxels 
and not yet mapped to $V_{inspected}$ (blue voxels in Fig. \ref{detect}(b)). 
To reduce computational complexity, exploration frontiers are clustered via K-means and inspection frontiers are grouped using region-growing algorithms.

\begin{figure}[t]
    \centering
    \includegraphics[scale=0.23]{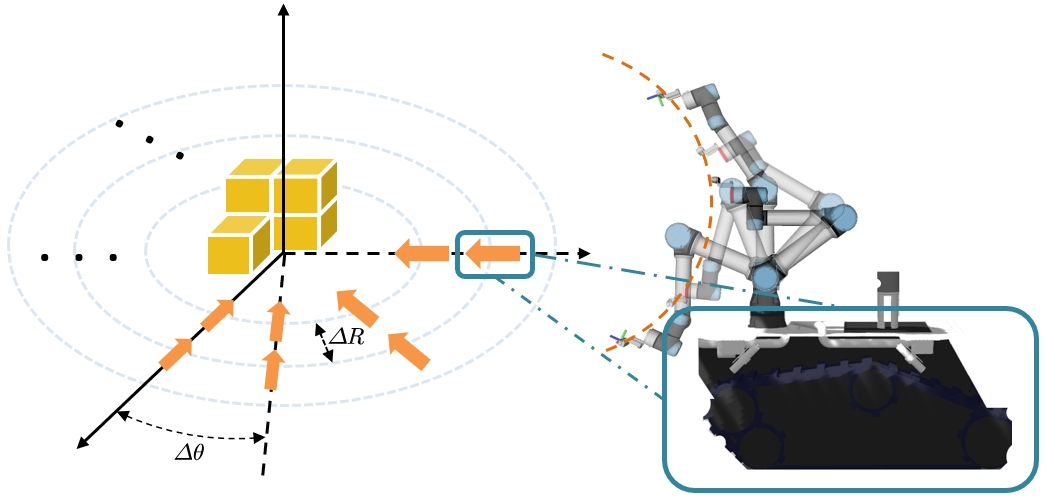}
    \caption{Generating viewpoint state library for a frontier cluster.
    The yellow arrows represent uniformly sampled mobile base poses, 
    combined with arm states to generate whole-body state of the mobile manipulator.
    $\varDelta R$ and $\varDelta \theta $ represent the radial resolution and angular resolution during sampling, respectively.
    Each whole-body state corresponds to a viewpoint.
    }
    \label{lib}
\end{figure}

\subsection{Viewpoints Generation and Evaluation}

Frontiers guide the robot in exploring unknown space and inspecting environments. 
Therefore, for each frontier cluster, a rich set of viewpoints must be generated to ensure comprehensive coverage. 
Existing approaches typically lack variation in orientation for generated viewpoints, resulting in insufficient coverage of frontiers \cite{Star}\cite{frontier_3}. 
Moreover, for redundant systems like mobile manipulators, solving for the robot state corresponding to a viewpoint through inverse kinematics (IK) incurs substantial computational costs.
To address these challenges, we introduce a viewpoint state library for mobile manipulators composed of a differential drive base and an $L$-DoF manipulator. 
As illustrated in Fig. \ref{lib}, multiple positions and yaw angles are uniformly sampled on the circumcircle of each frontier cluster centroid, forming a candidate set of base states. 
For each sampled base state $\bm{x}_b = \begin{bmatrix} q_x,q_y,\phi \end{bmatrix} ^\top$, $2N_{s}+1$ viewpoints with varying orientations are predefined under kinematic and collision-free constraints. 
Specifically, this includes one level viewpoint, $N_{s}$ downward viewpoints, and $N_{s}$ upward viewpoints, ensuring comprehensive coverage.
The joint states $\bm{x}_a = \begin{bmatrix} q_1,\cdots,q_L \end{bmatrix} ^\top$ corresponding to each viewpoint are computed via IK 
and combined with the base state to form the whole-body state $\bm{x} = \begin{bmatrix} \bm{x}_b^\top,\bm{x}_a^\top \end{bmatrix} ^\top$.
All viewpoints and their corresponding whole-body states constitute the viewpoint state library for the frontier cluster.
Each frontier cluster has a corresponding viewpoint state library.
 
The viewpoint state library shifts the computationally intensive IK calculations to an offline phase, 
reducing the online computational load while ensuring comprehensive coverage. 
The whole-body states provided by the viewpoint state library facilitate efficient viewpoint evaluation.
Similar to \cite{AEP}, we compute the motion gain $G_{move}$ and information gain $G_{info}$ for each viewpoint, with their product defining the final viewpoint score $G$:
$$
G = G_{move} \cdot G_{info}. \eqno{(1)}
$$

The motion gain $G_{move}$ is computed based on the time $t$ required for the robot to move from its current whole-body state $\bm{x}^{n}$ to the viewpoint's whole-body state $\bm{x}^{v}$:
$$
t = \max \left\{ \frac{L(\bm{x}_{b}^{n} , \bm{x}_{b}^{v})}{v_{max}} , \| (\bm{x}_{a}^{n} - \bm{x}_{a}^{v}) \oslash {\bm{\dot{q}}_{max}} \| _\infty \right\}, \eqno{(2)}
$$
$$
G_{move} = \text{exp}( -\lambda_\text{m} \cdot t),  \eqno{(3)}
$$
where $v_{max}$ denotes the maximum speed of the base, and $L(\bm{x}_{b}^{n}, \bm{x}_{b}^{v})$ represents the Euclidean distance 
between the robot's current base position and the base position corresponding to the viewpoint.
The joint state of the manipulator is signified by $\bm{x}_{a}$ and $\bm{\dot{q}}_{max}$ is a vector composed of the maximum angular velocities of the $L$ joints of the manipulator. 
The symbol $\oslash$ denotes element-wise division of vectors and $\lambda_\text{m}$ represents the penalty weight for time $t$.

Let $\lambda_\text{e}$ and $\lambda_\text{i}$ serve as the weights for the exploration task and the inspection task. 
The information gain $G_{info}$ for each viewpoint is determined through a weighted function:
$$
G_{info} = \lambda_\text{e} \cdot \left| B_{e}^{lidar} \cup B_{e}^{cam} \right| + \lambda_\text{i} \cdot \left| B_{i}^{rgb} \right|,  \eqno{(4)}
$$
where $B_{e}^{lidar}$ and $B_{e}^{cam}$ represent the sets of exploration frontiers covered by the LiDAR and the depth camera, respectively,
while $B_{i}^{rgb}$ denotes the set of inspection frontiers covered by visual inspection. 
We first filter out viewpoints with scores below a certain threshold, and then select the highest-scoring viewpoints
in each viewpoint state library, forming a candidate viewpoint set $C$ for subsequent decision-making.

\subsection{Optimal Viewpoint Decision}

\begin{algorithm}[t]
    \caption{Dual-Stage Decision Module}
    \label{algorithm}
    \KwIn{$C \leftarrow $ candidate viewpoint set 

    $\bm{x}^{n} \leftarrow $ robot current whole-body state 
    
    $M \leftarrow $ segmented map}
    \KwOut{optimal viewpoint $vp_{t}$}

    Initialize $R_{c}$ = CurrentRegion($\bm{x}^{n} , M $);

    Initialize $R_{t}$ = RegionDecider($C , M, R_c$);

    \For{each ${vp} \in {C}$}{
        \eIf{$\text{SameRegion}(R_c , vp)$}
        {
            ${C}_{n}$.PushBack($vp$);
        }{

            ${C}_{o}$.PushBack($vp$);
        }
    }

    \eIf{${C}_{n}.empty()$}
    {
        $vp_{t}$ = MinDistance($R_{t} , {C}_{o}$);
    }{
        $vp_{t}$ = ViewpointDecider($\bm{x}^{n} , R_{t} ,{C}_{n}$);
    }

\end{algorithm}

\begin{figure}[t]
    \centering
    \includegraphics[scale=0.225]{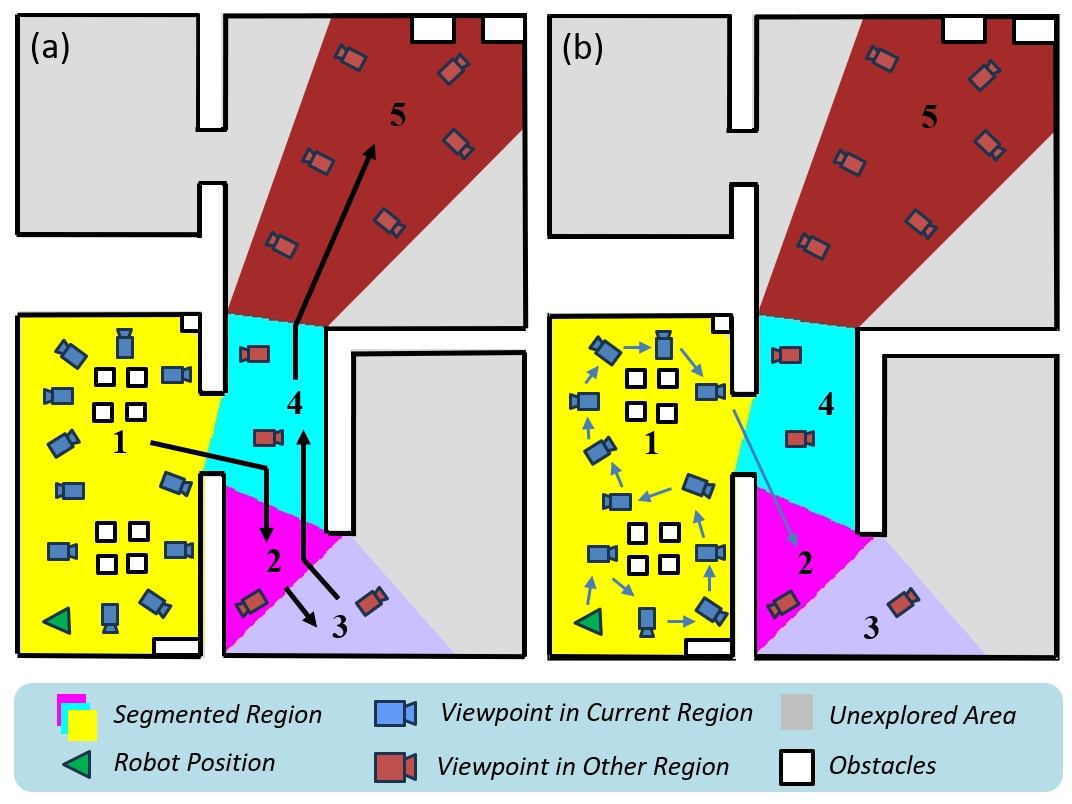}
    \caption{Results of the dual-stage decision module.
    (a) The region-level decision-making determines the visitation order of regions based on the partitioning results (black arrows and numbers).
    (b) The viewpoint-level decision-making determines the visitation order of viewpoints within the current region and proceeds to the target region (blue arrows).}
    \label{decider}
\end{figure}

Formulating an Asymmetric Traveling Salesman Problem (ATSP) can plan the shortest path to visit all viewpoints, avoiding local optima \cite{frontier_3}. 
However, the computational time grows significantly as the number of viewpoints increases.
To address this, we propose a dual-stage decision module, as outlined in Algorithm \ref{algorithm}, 
which determines the optimal viewpoint through region-level and viewpoint-level decisions in real-time. 
First, a contour-based segmentation algorithm is applied to the 2D cost map generated by the map manager \cite{segment}, 
partitioning it into the segmented map $M$ (colored areas in Fig. \ref{decider}(a)). Based on this partition, 
the CurrentRegion function (line 1) identifies the robot's current region $R_{c}$ (the yellow region in Fig. \ref{decider}(a)).
Next, the RegionDecider function (line 2) performs the region-level decision-making. 
Specifically, it uses the A* algorithm to compute the shortest paths between region centers, constructs an ATSP using these path lengths as traversal costs, 
and solves for the optimal visitation sequence starting from the current region. Once the region-level decision is complete, 
the first region in the sequence is selected as the target region $R_{t}$ (the purple region in Fig. \ref{decider}(a)).

Guided by $R_{t}$, the viewpoint-level decision-making (lines 3-13) determines the optimal viewpoint $vp_{t}$. 
Each candidate viewpoint in set $C$ is assigned to either $C_{n}$ (viewpoints within the robot's current region) 
or $C_{o}$ (viewpoints outside the current region). If ${C}_{n}$ is non-empty, it indicates that the current region has not been fully searched. 
The ViewpointDecider function (line 13) formulates an ATSP to plan the shortest path starting from the robot's current position, visiting all viewpoints in $C_{n}$,
and ending at the center of $R_{t}$ (Fig. \ref{decider}(b)). The first viewpoint along this path is selected as $vp_{t}$. 
Conversely, if ${C}_{n}$ is empty (indicating thorough coverage of the current region), the nearest viewpoint in $R_{t}$ is directly chosen as $vp_{t}$ (line 11).

The dual-stage decision module ensures real-time performance by filtering out viewpoints in irrelevant regions. 
It also prioritizes comprehensive coverage of the current region before transitioning to the other region, improving overall search efficiency.

\section{Whole-Body Motion Planner}

The proposed whole-body motion planner, illustrated in Fig. \ref{system}(b) and Fig. \ref{system}(c), 
includes a global path search module that combines Hybrid A* \cite{Hybrid} and Spatio-Temporal RRT* (ST-RRT*) to generate a safe and feasible initial whole-body path for the mobile manipulator. 
This initial path serves as the reference state and initial solution for the IPC optimization, 
ensuring rapid convergence of the Differential Dynamic Programming (DDP) algorithm while maintaining high-quality solutions.

\subsection{Global Path Search}

\begin{figure}[t]
    \centering
    \includegraphics[scale=0.21]{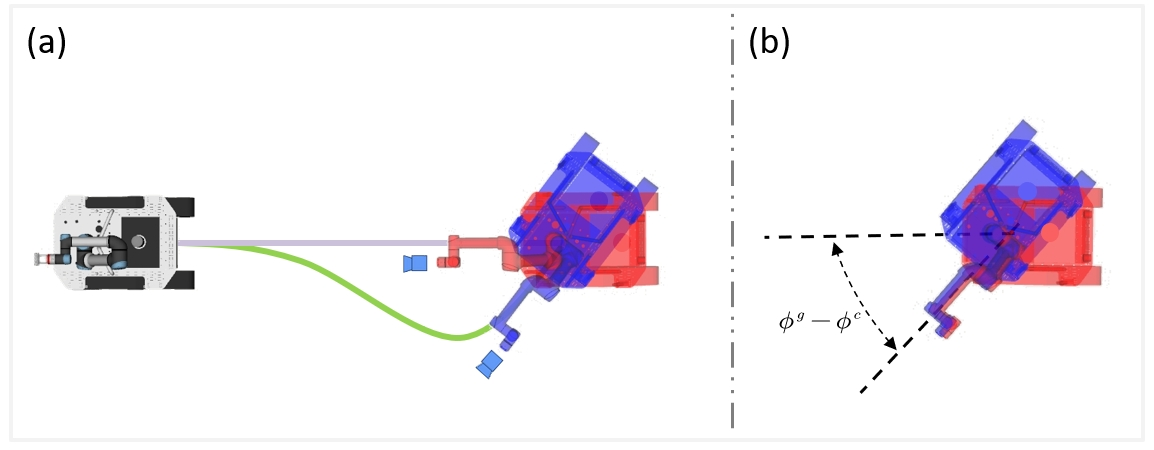}
    \caption{(a) The blue set represents the robot poses corresponding to the optimal viewpoint. Ignoring the target yaw angle $\phi^g$ during path planning(the red set) 
    can effectively reduce the mobile base path length (the purple line is shorter than the green line) but may impact the pose of the camera.
    (b) The camera pose is kept consistent by adjusting the joint angles of the target state, ensuring that the search performance is unaffected.}
    \label{edit}
\end{figure}

The viewpoint state library provides the motion planner with the whole-body state of the optimal viewpoint, 
which includes the robot's base state $\bm{x}_{b}^{g} = \begin{bmatrix} q_x^g,q_y^g,\phi^g \end{bmatrix} ^\top \in \text{SE}(2) $ 
and the manipulator state $\bm{x}_{a}^{g} \in \mathbb{R}^L$.
First, we use Hybrid A* to plan a path for the mobile base to reach $\bm{x}_{b}^{g}$. Given the 360° LiDAR's strong scene coverage capability, 
the target yaw angle $\phi^g$ is ignored during path planning. As shown in Fig. \ref{edit}(a), this approach significantly reduces path length 
but results in a discrepancy between the base's final orientation $\phi^c$ and the target yaw angle $\phi^g$, affecting the camera pose at the manipulator's end-effector.
To address this, the horizontal shoulder pan joint angle in $\bm{x}_{a}^{g}$ is adjusted by $\phi^g - \phi^c$, 
effectively compensating for the camera pose deviation (Fig. \ref{edit}(b)). This adjustment leverages the manipulator's mobility to compensate for the base's yaw discrepancy, 
improving search efficiency while minimizing energy consumption.
Finally, the planned base path is sampled at equal time intervals $\Delta t$ using a trapezoidal velocity profile, 
generating a sequence of $N$ waypoints $\mathbf{X}_B=\left\{\bm{x}_{b,1},\cdots,\bm{x}_{b,N}\right\}$ and their corresponding timestamps $\mathbf{T}=\left\{t_{1},\cdots,t_{N}\right\}$.

\begin{figure}[t]
    \centering
    \includegraphics[scale=0.285]{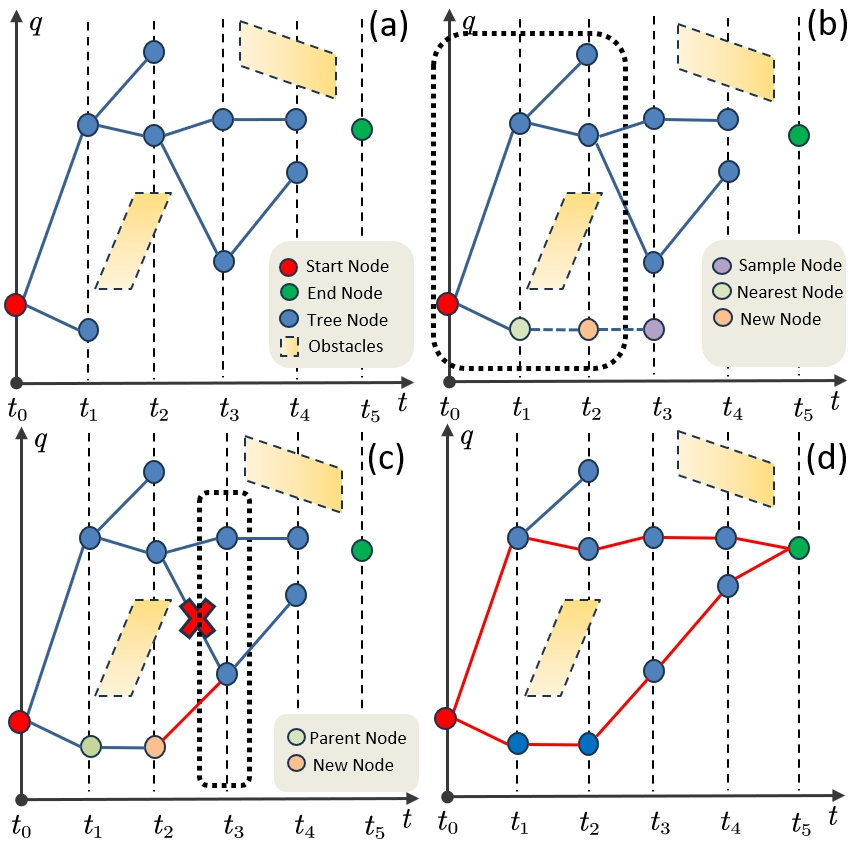}
    \caption{Example of ST-RRT*. The horizontal axis is the temporal configuration, and the vertical axe is the spatial configuration. 
    (a) ST-RRT* tree and two obstacles in S-T space. 
    (b) Illustration of the randomly sampled node $N_r$, the nearest node $N_n$, and the new node $N_{new}$ in ST-RRT*. 
    (c) Find a parent node for the new node. Delete the original edge (red cross) and rewire it to the new node to reduce the total path cost. 
    (d) Select the optimal solution from the red paths as the final solution of ST-RRT*.}
    \label{rrt}
\end{figure}

The proposed ST-RRT* extends the RRT* into the spatiotemporal dimensions, enabling the manipulator to safely and smoothly reach $\bm{x}_{a}^{g}$.
As illustrated in Fig. \ref{rrt}(a), each node in the search tree represents a feasible manipulator joint state associated with a specific timestamp $t_i$ in $\mathbf{T}$ (denoted by dashed lines).
Each manipulator state can be associated with a corresponding mobile base state based on timestamp, generating a whole-body state for collision detection.
The algorithm iteratively expands and optimizes the search tree until termination conditions are met.
Fig. \ref{rrt}(b) demonstrates the tree expansion process. A random node $N_{r}=\left\{\bm{x}_{a}^{r}, t_{r}\right\}$ is sampled 
from both the manipulator's feasible state space and temporal sequence $\mathbf{T}$. Subsequently, 
the nearest neighbor node $N_{n}=\left\{\bm{x}_{a}^{n},t_{n}\right\}$ is selected from candidates with timestamps preceding $t_{r}$ (nodes within dashed boxes). 
Through linear interpolation between $N_{n}$ and $N_{r}$, a new kinematically-constrained node $N_{new}=\left\{\bm{x}_{a}^{new},t_{n+1}\right\}$ is generated at the subsequent time step of $N_{n}$:
$$
\bm{x}_{a}^{new} = \bm{x}_{a}^{n} + \Delta t \cdot \min( \max(\frac{\bm{x}_{a}^{r} - \bm{x}_{a}^{n}}{t_{r} - t_{n}}, -\bm{\dot{q}}_{max} ) , \bm{\dot{q}}_{max} ) .\eqno{(5)}
$$

The optimization process of the search tree is illustrated in Fig. \ref{rrt}(c). 
We define the cost between nodes as the Euclidean distance between manipulator joint states to enhance motion smoothness.
ST-RRT* first identifies the feasible parent node with the lowest cost for $N_{new}$. Subsequently, 
a rewire operation is performed on the neighboring nodes of $N_{new}$ (nodes within the dashed box) to attempt connecting them to lower-cost paths, thereby optimizing the search tree structure.
Upon satisfying the termination conditions, as shown in Fig. \ref{rrt}(d), 
the terminal node is connected to the node at the previous timestep. 
The path that satisfies kinematic constraints and exhibits the minimum total cost is selected as the final result of ST-RRT*: $\mathbf{X}_A=\left\{\bm{x}_{a,1},\cdots,\bm{x}_{a,N}\right\}$.

By combining the mobile base state $\bm{x}_{b,i}$ from $\mathbf{X}_B$ with the corresponding manipulator state $\bm{x}_{a,i}$ from $\mathbf{X}_A$, 
the whole-body state of the mobile manipulator is generated as $\bm{x}_{i} = \begin{bmatrix} \bm{x}_{b,i}^\top,\bm{x}_{a,i}^\top \end{bmatrix} ^\top$.
This process ultimately forms the safe and feasible whole-body path of the mobile manipulator, denoted as $\mathbf{X}=\left\{\bm{x}_{1},\cdots,\bm{x}_{N}\right\}$.

\subsection{Local IPC Optimization}

After obtaining a high-quality mobile manipulator path from the global path search module, 
We formulate an IPC optimization problem to derive the optimal control strategy, presented as follows:
\begin{align*}
    \underset{\bm{x}, \bm{u}}{\arg\min} & \quad \varphi(\bm{x}(t_f)) + \int_{t_0}^{t_f}  L(\bm{x}(t),\bm{u}(t),t) dt \\
    \text{s.t.} & \quad \bm{x}(t_0) = \bm{x}_0, \\
    & \quad \dot{\bm{x}}(t) = {f}(\bm{x}(t), \bm{u}(t), t), \\
    & \quad {h}(\bm{x}(t), \bm{u}(t), t) > 0, \tag{6}
\end{align*} 
where $\bm{x}_0$ is the initial state,
$\dot{\bm{x}}(t) = {f}(\bm{x}(t), \bm{u}(t), t)$ is system kinematics model,
and $L(\bm{x}(t),\bm{u}(t), t)$ combines the whole-body state tracking error of the mobile manipulator and weighted control costs.
${h}(\bm{x}(t), \bm{u}(t), t) > 0$ is the inequality constraints that the robot must satisfy which include kinematic constraints (e.g., joint limits, maximum velocities) and obstacle avoidance constraints.

Linear inequalities are applied to bound the allowed joint angles as well as the velocity commands for the base and arm. 
Similar to \cite{DOG}, the mobile manipulator's geometry is approximated using a set of spheres for collision avoidance. 
By comparing each sphere's radius with the Euclidean Signed Distance Field (ESDF) value at its center, the following inequality constraint is constructed:
$$
D_{ESDF}\left(FK_{i}(\bm{x})\right) - r_i > 0 ,  \eqno{(7)}
$$
where $FK_{i}(\bm{x})$ represents the position of the $i$-th sphere's center computed through forward kinematics 
and $r_i$ denotes the radius of the $i$-th sphere. 
The function $D_{ESDF}(\cdot)$ retrieves the distance to the nearest obstacle from the ESDF. 
To ensure self-collision avoidance, the following inequality is constructed for any two spheres $i$ and $j$ in the set to prevent intersection:
$$
\|FK_{i}(\bm{x}) - FK_{j}(\bm{x})\| - r_i - r_j > 0. \eqno{(8)}
$$

We employ the Differential Dynamic Programming (DDP) method to solve the IPC problem in real-time.
Given that the iterative dynamic programming approach employed by DDP struggles to handle constraints directly, 
we introduce Relaxed Barrier Functions (RBF) \cite{RBF} to construct the cost function, penalizing violations of inequality constraints $z_i={h}(\bm{x}(t), \bm{u}(t), t) > 0$:
$$
B_i(z_i) = 
\begin{cases} 
-\mu \ln(z_i), & z_i > \delta \\
\mu ( -\ln(\delta) + (\frac{z_i - 2\delta}{\sqrt{2}\delta})^2 - 0.5 ). & z_i \leq \delta
\end{cases} \eqno{(9)}
$$
The parameters $\mu$ and $\delta$ can be used as tuning parameters for each constraint and we achieved good results with $\mu = 10^{-2}$ and $\delta = 10^{-3}$ for all constraints.
Ultimately, the optimal control strategy obtained by solving the IPC optimization problem drives the mobile manipulator to perform target search tasks 
with enhanced efficiency. 

\section{Simulations and Experiments}

\subsection{Implementation Details}

\begin{table}[t]
    \centering
    \caption{Experimental parameters.}
    \begin{tabular}{cccccccccc}
        \toprule
        $d_{ins}$ & $N_{s}$ & $\lambda_\text{m}$ & $\lambda_\text{e}$ & $\lambda_\text{i}$ &    $\varDelta R$    &   $\varDelta \theta $   &  $t_f$    \\
        \midrule
           4.0 m  &    3    &         0.5        &         0.2        &       0.8          &      0.3 m          &          30°            &   3.0 s  \\
        \bottomrule
    \end{tabular}
\end{table}

We build the IPC optimization framework based on the OCS2 toolbox \cite{OCS2}, 
employing an adaptive step-size ODE45 integrator for forward rollout integration. 
The construction method of the ESDF follows the approach described in \cite{frontier_3}. 
The ATSP problem in the dual-stage decision module is solved using the Lin-Kernighan-Helsgaun heuristic solver \cite{LKH}.
The parameters of our algorithm are listed in Table I. 

The simulation experiments utilize a mobile manipulator platform composed of a differential drive base and a UR5e manipulator.
The IPC optimization strategy solved in real-time directly controls the platform. 
For real-world experiments, a Robuster MR2000 differential drive base and an AUBO manipulator are employed, 
alongside an Intel RealSense D435i RGB-D camera and a Livox Mid-360 LiDAR.
An efficient LiDAR-inertial localization system \cite{FAST-LIO} and manipulator forward kinematics are used to estimate the poses of the mobile base and the end-effector.
All modules operate on an onboard Intel Core i5-9300H CPU without external dependencies.

\subsection{Benchmark and Analysis}

\begin{figure*}[t]
    \centering
    \includegraphics[scale=0.25]{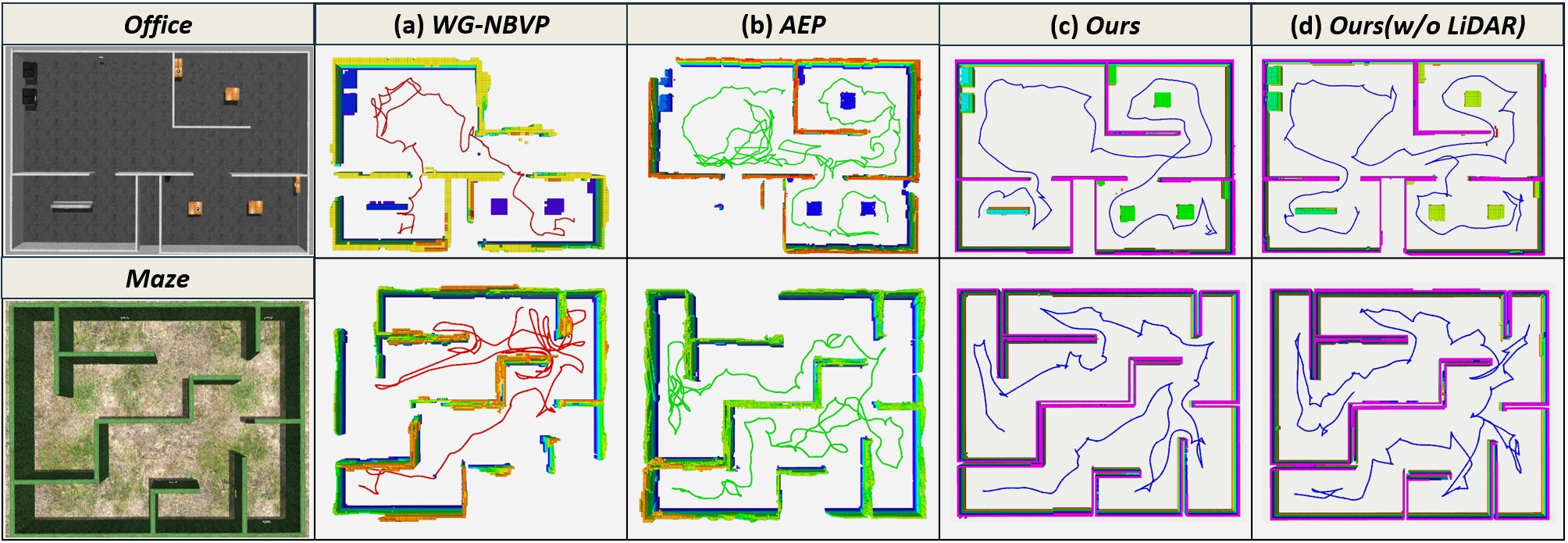}
    \caption{Trajectories of the mobile base executed by different methods in simulation experiments (office and maze).
    (a) The trajectories of WG-NBVP\cite{Sota} in two simulated environments.
    (b) The trajectories of AEP\cite{AEP}.
    (c) The trajectories of our method with LiDAR and RGB-D camera.
    (d) The trajectories generated by our method using only the depth camera mounted on the end-effector.}
    \label{exp}
\end{figure*}

\begin{table}[t]
    \centering
    \caption{Exploration statistic in the office and maze scenarios.}
    \begin{tabular}{ccccc}
        \toprule
        \multirow{2}{*}{\textbf{Scene}} & \multirow{2}{*}{\textbf{Method}} & \multicolumn{2}{c}{\textbf{Search time(s)}} & {\textbf{Complet}} \\
                               &                         & \textbf{Avg}       & \textbf{Std}     &\textbf{-eness(\%)}                                   \\
        \midrule
        Office & WG-NBVP\cite{Sota} & $>$900 & N/A & 42.5 \\
               & AEP\cite{AEP} & $>$900 & N/A & 50.0 \\
               & Ours(w/o LiDAR) & 581 & 25 & 100.0 \\
               & Ours & \textbf{556} & \textbf{13} & \textbf{100.0} \\
        \midrule
        Maze   & WG-NBVP\cite{Sota} & $>$900 & N/A & 60.0 \\
               & AEP\cite{AEP} & $>$900 & N/A & 85.0 \\
               & Ours(w/o LiDAR) & 468 & 27 & 100.0 \\
               & Ours & \textbf{420} & \textbf{21} & \textbf{100.0} \\
        \bottomrule
    \end{tabular}
\end{table}

\begin{table}[t]
    \centering
    \caption{Average path length and joint rotation angle of each method.}
    \begin{tabular}{cccc}
        \toprule
        \multirow{2}{*}{\textbf{Scene}} & \multirow{2}{*}{\textbf{Method}} & {\textbf{Path}} & {\textbf{Total Joint }}  \\
                                        &                                  &   {\textbf{Length(m)}}    &  {\textbf{Rotation Angle(rad)}} \\
        \midrule
        Office & WG-NBVP\cite{Sota} & 154.81          &  750.35\\
               & AEP\cite{AEP}      & 198.28          &  1102.24\\
               & Ours(w/o LiDAR)    & 149.35          &  391.11\\
               & Ours               & \textbf{140.63} &  \textbf{383.23}\\
        \midrule
        Maze   & WG-NBVP\cite{Sota} & 158.13          &  910.28\\
               & AEP\cite{AEP}      & 165.40          &  1304.68\\
               & Ours(w/o LiDAR)    & 152.04          &  428.14\\
               & Ours               & \textbf{143.72} &  \textbf{415.63}\\
        \bottomrule
    \end{tabular}
\end{table}

We construct an office scene (30 m $\times$ 20 m $\times$ 3 m) and a maze scene (25 m $\times$ 22 m $\times$ 3 m) in Gazebo for simulation experiments. 
Each scene includes eight search targets (represented by apriltags), placed on walls, cabinets, trash bins, and under tables to comprehensively 
evaluate the search completeness of different methods. We compare the proposed framework with WG-NBVP \cite{Sota} (state-of-the-art) 
and AEP \cite{AEP} (classic). Considering that WG-NBVP \cite{Sota} and AEP \cite{AEP} only use the depth camera at the end-effector 
for perception, an ablation experiment is conducted where our method also uses only the end-effector depth camera. 
In all tests, the voxel resolution is set to 0.15 m, the maximum speed and maximum angular velocity of the mobile base 
are set to 0.8 m/s and 1.5 rad/s, and the maximum angular velocity of the manipulator joints is set to 0.8 rad/s. 
Each method is run 10 times in both scenes with the same initial configuration, with a time limit of 15 minutes per run. 
We record the efficiency (search time), search completeness (percentage of detected apriltags), the path length of the mobile base, 
and total rotation angle of the manipulator's joints for the four methods. The results are shown in Tables II and III, 
and the trajectories of the mobile base during the target search are illustrated in Fig. \ref{exp}.

The results indicate that, due to the lack of a rational viewpoint planner and an efficient motion planning method, 
WG-NBVP \cite{Sota} and AEP \cite{AEP} struggle to achieve comprehensive search within the 15-minute test duration. 
Our target viewpoint planner not only efficiently accomplishes exploration tasks but also leverages its high DoF to inspect visually occluded objects, 
seamlessly integrating exploration with inspection.
Additionally, the coordinated operation of the target viewpoint planner and the whole-body motion planner achieves higher search coverage 
with reduced path length and rotation angle, further enhancing target search efficiency.
Compared to using a single sensor, the combination of a depth camera and LiDAR yields optimal results in target search tasks.
The LiDAR's wide-range perception improves exploration efficiency and provides more information for viewpoint decision-making, 
while the depth camera on the manipulator's end-effector effectively covers the LiDAR's blind spots. 
The complementary advantages of these heterogeneous sensors ensure both efficiency and completeness in target search.
 
\begin{figure}[t]
    \centering
    \includegraphics[scale=0.33]{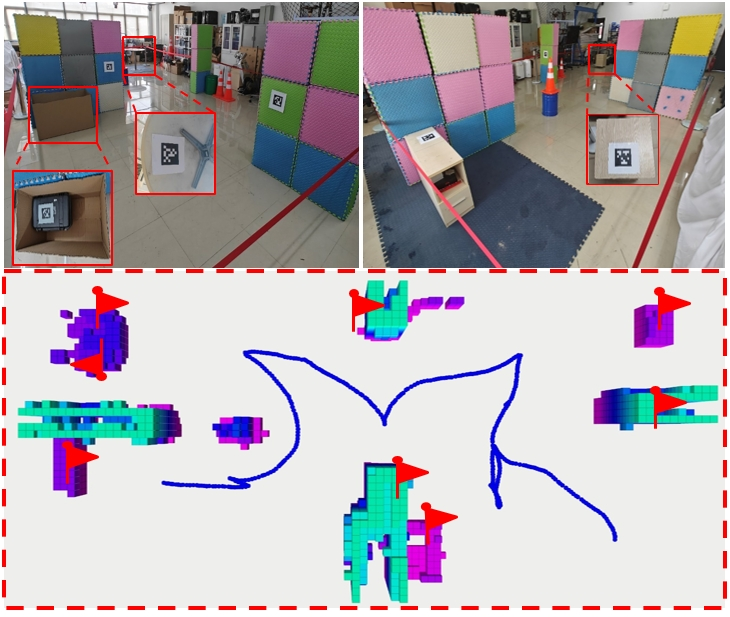}
    \caption{Experiments conducted in a indoor environment with 8 apriltags to be found. 
    The target locations include inside the box, under the table, and on top of the cabinet.
    The trajectory of our method is shown in the figure, where the red flags represent the successfully located apriltags.}
    \label{real}
\end{figure}

\subsection{Real-world Experiments}

We validate the proposed method in real-world experiments.
The experimental setup is an indoor area measuring 12 m $\times$ 5 m $\times$ 3 m, enclosed by isolation belts.
Eight apriltags are placed at various locations as search targets to evaluate the capabilities of our approach. 
The maximum speed and maximum angular velocity of the mobile base are set to 0.5 m/s and 1.2 rad/s, 
while the maximum joint angular velocity of the manipulator is set to 0.6 rad/s. The maximum observation distance 
$d_{ins}$ and voxel resolution are configured to 3.0 m and 0.1 m. 
Other algorithms struggle to find targets in specific positions (inside boxes, under tables, and on cabinets),
and the decoupled motion planning approach results in unnecessary motion cost.
The proposed method completes the search for all targets within 118 seconds 
while ensuring the safety and feasibility of the mobile manipulator's whole-body trajectory. 
The experimental scene and the online-generated map are shown in Fig. \ref{real}. 
The path length of the mobile base and the total rotation angle of the manipulator joints are 16.86 m and 52.74 rad. 
This experiment demonstrates the algorithm's capability in complex real-world scenarios. 
Further details can be found in the video demonstration.

\section{CONCLUSIONS}

In this paper, we propose a hierarchical framework designed for mobile manipulators to enhance their target search capabilities 
in complex and unknown environments. We introduce a target viewpoint planner, which achieves comprehensive coverage of exploration 
and inspection frontiers through a viewpoint state library and a well-designed evaluation function. 
A dual-stage decision module significantly reduces detours and repeated visits. 
Furthermore, a whole-body motion planner enables the robot to quickly and safely visit various viewpoints, improving search efficiency. 
Both simulation and real-world experiments demonstrate the effectiveness of our approach.

In future work, we plan to incorporate Vision-Language Models (VLM) to enhance the robot's understanding of the environment 
and to reasonably assess the likelihood of target presence in different locations. By leveraging an end-to-end model, 
the robot can prioritize areas with higher probabilities of containing targets, thereby further improving search efficiency.

\addtolength{\textheight}{-12cm}   




\bibliographystyle{ieeetr}
\bibliography{reference}

\begin{thebibliography}{10}

\bibitem{9570974}
H.~Wu, L.~Jiang, X.~Liu, J.~Li, Y.~Yang, and S.~Zhang, ``Intelligent explosive ordnance disposal uav system based on manipulator and real-time object detection,'' in {\em 2021 4th International Conference on Intelligent Robotics and Control Engineering (IRCE)}, pp.~61--65, 2021.

\bibitem{Wirth_Pellenz_2007}
S.~Wirth and J.~Pellenz, ``Exploration transform: A stable exploring algorithm for robots in rescue environments,'' in {\em 2007 IEEE International Workshop on Safety, Security and Rescue Robotics}, p.~1–5, Sep 2007.

\bibitem{Yokokohji_2021}
Y.~Yokokohji, ``The use of robots to respond to nuclear accidents: Applying the lessons of the past to the fukushima daiichi nuclear power station,'' {\em Annual Review of Control, Robotics, and Autonomous Systems}, p.~681–710, May 2021.

\bibitem{10166021}
X.~Li, X.~Li, S.~Gu, and J.~Li, ``A cooperative target search and rescue guidance system design based on acoustic spectrum analysis,'' in {\em 2023 4th International Conference on Computer Vision, Image and Deep Learning (CVIDL)}, pp.~541--544, 2023.

\bibitem{10451197}
G.~Deng, X.~Yao, B.~Wang, X.~He, and Q.~Fei, ``Research on uav coverage search based on ddqn in unknown environments,'' in {\em 2023 China Automation Congress (CAC)}, pp.~2826--2831, 2023.

\bibitem{Star}
Y.~Luo, Z.~Zhuang, N.~Pan, C.~Feng, S.~Shen, F.~Gao, H.~Cheng, and B.~Zhou, ``Star-searcher: A complete and efficient aerial system for autonomous target search in complex unknown environments,'' {\em IEEE Robotics and Automation Letters}, vol.~9, no.~5, pp.~4329--4336, 2024.

\bibitem{frontier_1}
B.~Yamauchi, ``A frontier-based approach for autonomous exploration,'' in {\em Proceedings 1997 IEEE International Symposium on Computational Intelligence in Robotics and Automation CIRA’97. “Towards New Computational Principles for Robotics and Automation”}, Nov 2002.

\bibitem{frontier_2}
D.~Deng, R.~Duan, J.~Liu, K.~Sheng, and K.~Shimada, ``Robotic exploration of unknown 2d environment using a frontier-based automatic-differentiable information gain measure,'' in {\em 2020 IEEE/ASME International Conference on Advanced Intelligent Mechatronics (AIM)}, pp.~1497--1503, 2020.

\bibitem{frontier_3}
B.~Zhou, Y.~Zhang, X.~Chen, and S.~Shen, ``Fuel: Fast uav exploration using incremental frontier structure and hierarchical planning,'' {\em IEEE Robotics and Automation Letters}, p.~779–786, Apr 2021.

\bibitem{tare}
C.~Cao, H.~Zhu, H.~Choset, and J.~Zhang, ``Tare: A hierarchical framework for efficiently exploring complex 3d environments,'' in {\em Robotics: Science and Systems XVII}, Jun 2021.

\bibitem{Schmid_Pantic_Khanna_Ott_Siegwart_Nieto_2020}
L.~Schmid, M.~Pantic, R.~Khanna, L.~Ott, R.~Siegwart, and J.~Nieto, ``An efficient sampling-based method for online informative path planning in unknown environments,'' {\em IEEE Robotics and Automation Letters}, p.~1500–1507, Apr 2020.

\bibitem{sample_2}
A.~Bircher, M.~Kamel, K.~Alexis, H.~Oleynikova, and R.~Siegwart, ``Receding horizon “next-best-view” planner for 3d exploration,'' in {\em 2016 IEEE International Conference on Robotics and Automation (ICRA)}, May 2016.

\bibitem{COME1}
B.~Charrow, G.~Kahn, S.~Patil, S.~Liu, K.~Goldberg, P.~Abbeel, N.~Michael, and V.~Kumar, ``Information-theoretic planning with trajectory optimization for dense 3d mapping,'' in {\em Robotics: Science and Systems XI}, Jan 2016.

\bibitem{Semantic}
S.~Papatheodorou, N.~Funk, D.~Tzoumanikas, C.~Choi, B.~Xu, and S.~Leutenegger, ``Finding things in the unknown: Semantic object-centric exploration with an mav,'' in {\em 2023 IEEE International Conference on Robotics and Automation (ICRA)}, pp.~3339--3345, 2023.

\bibitem{Trolley}
J.~Gao, P.~Xie, X.~Gao, Z.~Sun, J.~Wang, and M.~Q.-H. Meng, ``Indoor exploration and simultaneous trolley collection through task-oriented environment partitioning,'' in {\em 2024 IEEE International Conference on Robotics and Automation (ICRA)}, pp.~14895--14901, 2024.

\bibitem{Pin}
S.~Isler, R.~Sabzevari, J.~Delmerico, and D.~Scaramuzza, ``An information gain formulation for active volumetric 3d reconstruction,'' in {\em 2016 IEEE International Conference on Robotics and Automation (ICRA)}, May 2016.

\bibitem{Sota}
M.~Naazare, F.~G. Rosas, and D.~Schulz, ``Online next-best-view planner for 3d-exploration and inspection with a mobile manipulator robot,'' {\em IEEE Robotics and Automation Letters}, vol.~7, no.~2, pp.~3779--3786, 2022.

\bibitem{AEP}
M.~Selin, M.~Tiger, D.~Duberg, F.~Heintz, and P.~Jensfelt, ``Efficient autonomous exploration planning of large-scale 3-d environments,'' {\em IEEE Robotics and Automation Letters}, p.~1699–1706, Apr 2019.

\bibitem{segment}
L.~Fermin-Leon, J.~Neira, and J.~A. Castellanos, ``Incremental contour-based topological segmentation for robot exploration,'' in {\em 2017 IEEE International Conference on Robotics and Automation (ICRA)}, p.~2554–2561, May 2017.

\bibitem{Hybrid}
D.~Dolgov, S.~Thrun, M.~Montemerlo, and J.~Diebel, ``Path planning for autonomous vehicles in unknown semi-structured environments,'' {\em The International Journal of Robotics Research}, p.~485–501, Apr 2010.

\bibitem{DOG}
J.-R. Chiu, J.-P. Sleiman, M.~Mittal, F.~Farshidian, and M.~Hutter, ``A collision-free mpc for whole-body dynamic locomotion and manipulation,'' in {\em 2022 International Conference on Robotics and Automation (ICRA)}, pp.~4686--4693, 2022.

\bibitem{RBF}
C.~Feller and C.~Ebenbauer, ``Relaxed logarithmic barrier function based model predictive control of linear systems,'' {\em IEEE Transactions on Automatic Control}, vol.~62, no.~3, pp.~1223--1238, 2017.

\bibitem{OCS2}
F.~Farshidian {\em et~al.}, ``{OCS2}: An open source library for optimal control of switched systems.''
\newblock [Online]. Available: \url{https://github.com/leggedrobotics/ocs2}.

\bibitem{LKH}
K.~Helsgaun, ``An effective implementation of the lin–kernighan traveling salesman heuristic,'' {\em European Journal of Operational Research}, p.~106–130, Oct 2000.

\bibitem{FAST-LIO}
W.~Xu and F.~Zhang, ``Fast-lio: A fast, robust lidar-inertial odometry package by tightly-coupled iterated kalman filter,'' {\em IEEE Robotics and Automation Letters}, vol.~6, no.~2, pp.~3317--3324, 2021.

\end{thebibliography}

\end{document}